\documentclass{article}

\usepackage{microtype}
\usepackage{graphicx}
\usepackage{subfigure}
\usepackage{booktabs} % for professional tables

\usepackage{times}
\usepackage{epsfig}
\usepackage{amsmath}
\usepackage{amssymb}
\usepackage{multirow}
\usepackage{array}
\usepackage{makecell}
\usepackage{bm}
\usepackage{enumerate}
\usepackage[T1]{fontenc}
\usepackage{algorithm}
\usepackage{algorithmic}
\usepackage[breaklinks=true,bookmarks=false]{hyperref}

\usepackage[preprint]{neurips_2020}

\title{Credit Assignment with Meta-Policy Gradient for Multi-Agent Reinforcement Learning}

\author{
Jianzhun Shao, Hongchang Zhang, Yuhang Jiang, Shuncheng He, Xiangyang Ji\\
Department of Automation\\
Tsinghua University\\
\{sjz18,hc-zhang19,jiangyh19,hesc16\}@mails.tsinghua.edu.cn, xyji@tsinghua.edu.cn
}

\begin{document}
\maketitle
\begin{abstract}
Reward decomposition is a critical problem in centralized training with decentralized execution~(CTDE) paradigm for multi-agent reinforcement learning. To take full advantage of global information, which exploits the states from all agents and the related environment for decomposing Q values into individual credits, we propose a general meta-learning-based Mixing Network with Meta Policy Gradient~(MNMPG) framework to distill the global hierarchy for delicate reward decomposition. The excitation signal for learning global hierarchy is deduced from the episode reward difference between before and after ``exercise updates'' through the utility network. Our method is generally applicable to the CTDE method using a monotonic mixing network. Experiments on the StarCraft II micromanagement benchmark demonstrate that our method just with a simple utility network is able to outperform the current state-of-the-art MARL algorithms on 4 of 5 super hard scenarios. Better performance can be further achieved when combined with a role-based utility network.
% When just adopting the simple utility network, our method outperforms the current state-of-the-art MARL algorithms on 4 of 5 super hard scenarios, while combined with role-based utility network, our method has further better performance and outperforms all 5 super hard scenarios and other hard scenarios.
\end{abstract}

\section{Introduction}
Multi-agent deep reinforcement learning algorithms (MARL) have recently shown extraordinary performance in various games like DOTA2~\cite{berner2019dota}, StarCraft~\cite{samvelyan2019starcraft}, and Honor of Kings~\cite{ye2020towards}. The framework of centralized training with decentralized execution (CTDE)~\cite{gupta2017cooperative,rashid2018qmix}, which enjoys the advantages of joint action learning~\cite{claus1998dynamics} and independent learning~\cite{tan1993multi}, is one of the popular frameworks for solving collaborative multi-agent tasks. 

Recent research on the CTDE framework can be divided into two categories: One is to enhance agents' ability with emphasis on individually processing local observations, for instance, allocating a role~\cite{wang2020roma, wang2020rode} or a mode of exploration~\cite{mahajan2019maven} to each agent, which may need extra prior information and consume more computational resources on decentralized execution due to extensive inference network. Another aims to decompose the single reward to each agent accurately, in other words, training a delicate mixing network for local utility values~\cite{rashid2018qmix,yang2020qatten,wang2020qplex}, or training a new joint action-value to factorizing tasks~\cite{son2019qtran}. The latter's empirical performance on challenging tasks is relatively limited by training instability and demand for delicate parameter adjustment. 

The current methods on mixing network usually utilize the global information from the full state roughly, such as ~\cite{rashid2018qmix} which only takes the full state vector as the input of the mixing network. These methods lack an explicit hierarchy, the distilled information from the full state, to decompose Q values into individual credit. Inspired by Meta-DDPG~\cite{xu2018learning}, where meta-learning is adopted for exploration, in this paper, we present a general meta-learning-based framework called Mixing Network with Meta Policy Gradient~(MNMPG) for exploration on better reward decomposition. Different from current role-based methods, we consider learning better Q combination ways through meta policy gradient, that is, a ``role'' conditioned on the global state, trained to guide a more reasonable overall $Q$ value, rather than a role conditioned on each agent's local observation. 

Our insight is that we expect to seize a hierarchy, representing some exploration inclination that may produce a higher future reward. Specifically, with meta-learning, we first collect trajectories with the initial utility policy. Then we perform ``exercise move'' by several Q-learning updates under the guidance of the global hierarchy. Next, we utilize the updated utility policy to generate new trajectories. The reward difference between the episode reward before and after ``exercise move'' plays as the excitation signal for finally training the global hierarchy. To some extent, the reward difference represents the ability to make better exploration for a global hierarchy. Since meta-learning is entirely based on self-improvement, no global priors are required. 

As an add-on of the procedure of the mixing network, MNMPG is generally applicable to many CTDE methods like QMIX~\cite{rashid2018qmix}, ROMA~\cite{wang2020roma}, RODE~\cite{wang2020rode}, MAVEN~\cite{mahajan2019maven}, etc.
Experiments on SMAC~\cite{samvelyan2019starcraft} show that our method with a simple 2-layer utility network from QMIX exceeds the current state-of-the-art on four of five \textbf{super hard} maps. Ablations demonstrate the remarkable superiority of our exploration strategy. In addition, we implement MNMPG based on state-of-the-art method RODE, also achieving better and more stable performance.

\section{Background}
In our work, we consider a fully collaborative multi-agent task with $n$ agents, which can be modeled as a \emph{decentralised
partially observable Markov decision process} (Dec-POMDP)~\cite{oliehoek2016concise} $G=\langle S,A,I,P,r,Z,O,n,\gamma \rangle$, where $s\in S$ is the true state of the environment. At each time step $t$, each agent $i\in I\equiv\{1,...,n\}$ chooses an action $a_i\in A$, forming a joint action $\mathbf{a}\in \mathbf{A}\equiv A^n$. $P(s_{t+1}|s_t, \mathbf{a}_t):S\times\mathbf{A}\times S \rightarrow [0,1]$ is the state transition function of the environment. All agents share the same reward function $r(s, \mathbf{a}):S\times\mathbf{A}\rightarrow \mathbb{R}$. $\gamma \in [0, 1)$ is the discount factor.
Due to the \emph{partial observability}, each agent $i$ has its local observations $z\in Z$ drawn from the observation function $O(s, i):S\times I \rightarrow Z$. Each agent chooses an action by its stochastic policy $\pi^i (a_i|\tau_i):A\times T\rightarrow [0,1]$, where $\tau_i\in T\equiv(Z\times A)^*$ denotes the action-observation history of agent $i$, and $\mathbf{\tau}$ is the action-observation histories of all agents. The agents’ joint policy $\pi$ induces a joint \emph{action-value function}: $Q^{\pi}(s_t, \mathbf{a}_t)=\mathbb{E}_{s_{t+1:\infty}, \mathbf{a}_{t+1:\infty}}[R_t|s_t,\mathbf{a}_t]$, where $R_t=\sum^\infty_{k=0}\gamma^k r_{t+k}$ is the discounted accumulated reward. The goal of our method is to find the optimal joint policy $\pi^*$ such that $Q^{\pi^*}(s,\mathbf{a})\ge Q^{\pi}(s, \mathbf{a})$, for all $\pi$ and $(s,\mathbf{a})\in S\times \mathbf{A}$.

A straightforward way to solve the problem is joint action learning~\cite{claus1998dynamics}, which utilizes a centralized policy to process the full state information and gives actions to all agents. This way is hard for distributed deployment and inefficient when the number of agents is large. Another way is independent learning~\cite{tan1993multi}, which regards other agents as part of the environment. It suffers from non-stationarity since other agents' performance changes as well during training. A solution combining the advantages of such two methods is the framework of centralized training with decentralized execution (CTDE)~\cite{gupta2017cooperative,rashid2018qmix}. During training, the method has access to the full state $s$ and every agent's action-observation history. As a result, information flow between all agents is allowed. During testing(execution), each agent only has access to its own action-observation history. 

Reward decomposition is a promising way to exploit the CTDE paradigm~\cite{rashid2018qmix,foerster2018counterfactual,mahajan2019maven,wang2020roma,rashid2020weighted}, which is a composite of a local utility network for each agent's execution and a mixing network for combining local utilities into a global action value. Many existing methods try to learn a compelling local utility network, which typically needs a relatively extensive network and more execution time. In this paper, we show that a simple local utility network combined with a delicate mixing network can achieve superior performance.
\begin{figure*}[!htbp]
\begin{center}
\includegraphics[width=\textwidth]{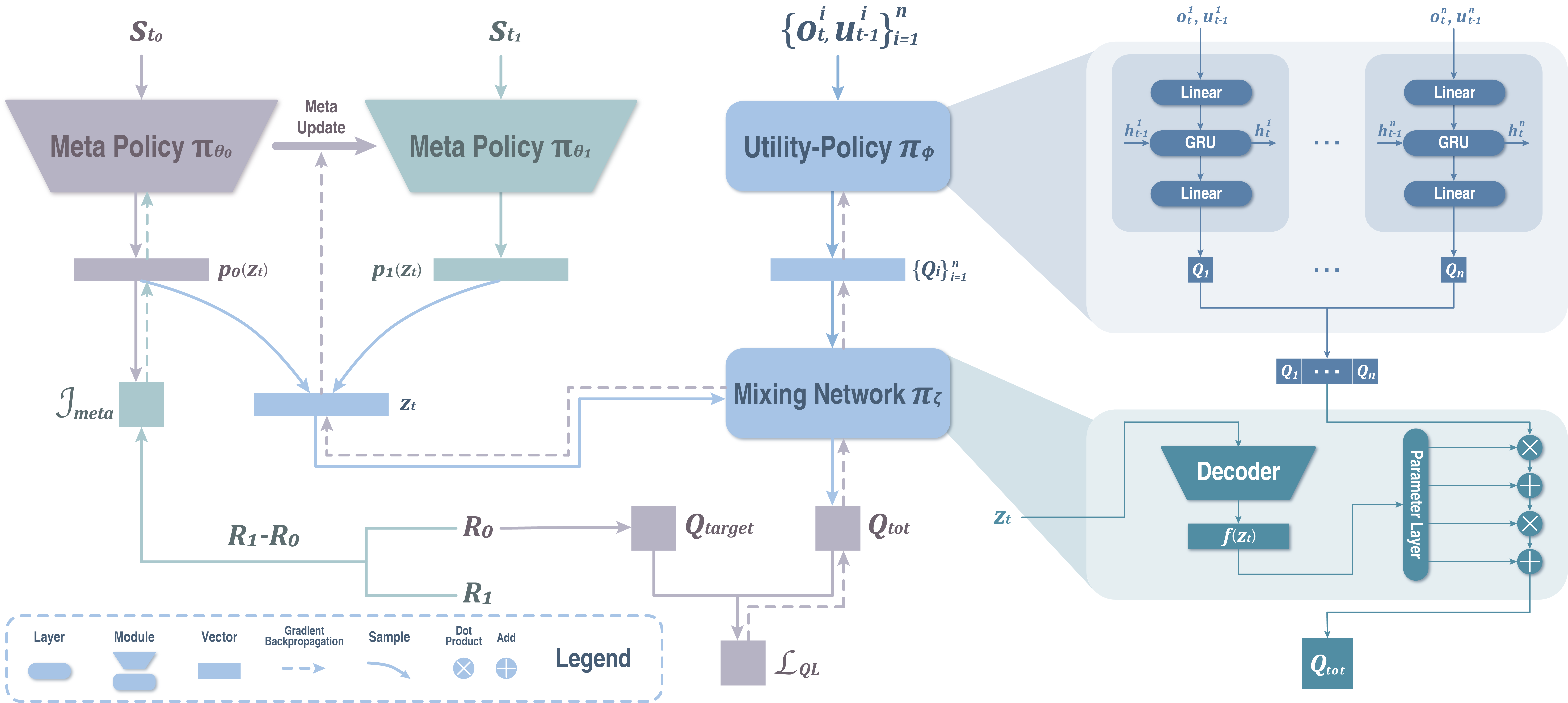}
\end{center}
\caption{The MNMPG framework.}
\label{framework}
\end{figure*}
\section{Method}
In this section, we first introduce the framework of MNMPG, and then present the update rules and the whole training procedure. Finally, the adaptation of MNMPG is presented.
\subsection{Framework}
Fig. \ref{framework} shows the MNMPG framework in terms of the CTDE paradigm. It learns a local Q value functions $Q^{\pi}_i$ for each agent, fed into a mixing network to compute a global TD loss for centralized training. During execution, only the local Q value functions are executed, while the mixing network is discarded. Therefore, each agent's acting policy is deducted from its local observations.

In the centralized training part, we propose to estimate the global Q value function $Q_{tot}$ with the guidance of a global hierarchy $z_t$. We divide the mixing network into two parts: one to infer the global hierarchy $z_t$, the other to estimate $Q_{tot}$. $Q_{tot}$ is parameterized by $\zeta$:
\begin{equation}
    Q_{tot}(\bm{\tau}, \mathbf{a}) := f_{\zeta}(Q_i(\tau^i, a^i_t)|z_t),  \ \ \ \ \ \ \ i\in [1,n].
\end{equation}

It's worth noting that our method differs from the traditional way to infer $Q_{tot}$ which conditions it on the full state $s_t$~\cite{rashid2018qmix} via $Q_{tot}(\bm{\tau}, \mathbf{a}) := f_{\zeta}(Q_i(\tau^i, a^i_t)|s_t), \ i\in [1,n]$. Another kind of methods that have a similar form to ours is the role-based utility network, which embeds the local Q values $Q_i$ with a role or a mode of exploration~\cite{mahajan2019maven,wang2020roma,wang2020rode} via
$Q_{tot}(\bm{\tau}, \mathbf{a}) := f_{\zeta}(Q_i(\tau^i, a^i_t|z_i)|s_t),  \ \ \ \ \ \ \ i\in [1,n].$
The reason lies in:
\begin{enumerate}[i)]
    \item It lacks a clear global objective by merely using the full state as global guidance for local Q functions. When players play StarCraft, they usually have a small goal for each game stage, such as ``eliminate Zergling 2'' or ``move backward''. The same local Q value for such different situations may contribute different weights for $Q_{tot}$. A brief hierarchy can better reflect high-level goals than the full state for mixing network to learn. A global hierarchy can also lead to better collaborative exploration.
    \item Traditional monotonic mixing network has fundamental flaws to solve specific tasks, which has been clearly stated by ~\cite{mahajan2019maven,son2019qtran,rashid2020weighted}. We can alleviate the problem by introducing a nonmonotonic global hierarchy to control the weight each local Q value possesses in $Q_{tot}$.
    \item Learning hierarchy for mixing network rather than utility network is more efficient for both training and execution. We do not need to assign each agent a role for each time step, namely only one temporary overall mode is learned. Since the hierarchy only affects the mixing network, which is removed during execution, we can perform executions efficiently with a small network.
\end{enumerate}
At each time step, the hierarchy $z_t$ is sampled from a joint Gaussian Distribution of $K$ dimension, whose mean and variance are produced by a neural network $f_{\theta}$, taking the full state as input:
\begin{equation}
z_t \sim \mathcal{N}(\mu_{\theta}(s_t), \sigma_{\theta}(s_t)).
\end{equation}
In section \ref{upr}, we will depict the details of learning hierarchy $z_t$ and the general update rules. After getting the hierarchy $z_t$, it is fed into a decoder, and outputs of the decoder $g_{\zeta}(z_t)$ are restricted to be non-negative. Similar to ~\cite{rashid2018qmix}, rather than concatenating the decoded vector to the local $Q_i$, we expect to be less constraining. We hope the global hierarchy dominates the weights of local value functions arbitrarily to explore the most suitable combination of local $Q_i$. Therefore the outputs of local utility network $Q_i$ are fed into a two-layer network and apply dot product with $g_{\zeta}(z_t)$ at each layer separately. The final output is $Q_{tot}$.
\subsection{Meta Policy Gradient for Mixing Network}\label{upr}
Our framework concentrates on learning a hierarchy to guide the combination of local Q values. This hierarchy should embody the properties: First, it is a time-varying hierarchy that representing the current goal of the team, so it is calculated from $s_t$ at each time step. It differs from \cite{mahajan2019maven}, who computes a ``goal'' from the initial state $s_0$, representing the exploration mode for the whole episode. Second, unlike \cite{wang2020rode} who adjusts the number of roles as a hyperparameter for each map, we prefer the hierarchy containing no prior information and can be improved spontaneously. In the following statement, we use $\pi_{\theta}$ to represent the policy that computes global hierarchy $z$, $\pi_{\zeta}$ for mixing network
, and $\pi_{\phi}$ for local utility network.

Inspired by \cite{xu2018learning}, who learns a noise policy for exploration with the episode reward difference, we propose to use meta-learning to learn a hierarchy without any additional prior information. At each iteration, we first generate a set of data $D_0$. Our goal is to adaptively improve the hierarchical network $\pi_{\theta}$  to combine the local Q values the most reasonably to make the utility network learner improve as fast as possible.

In this meta-framework, the generation of data $D_0$ can be viewed as the ``action'' taken by the hierarchical network, and its related reward should be defined as the improvement of the $Q_{tot}$ with Q-learning using data $D_0$,
\begin{equation}
    \begin{aligned}
    \mathcal{J}_{meta}(\pi_{\theta}) &=\mathbb{E}_{D_{0} \sim \pi_{\theta,\phi,\zeta}}\left[\mathcal{R}\left(\pi, D_{0}\right)\right] \\ &=\mathbb{E}_{D_{0} \sim \pi_{\theta,\phi,\zeta}}\left[R_{\pi^{\prime}}-R_{\pi}\right],
    \end{aligned}
\end{equation}
where $\pi^{\prime}=QL(\pi, D_0)$ denotes a new policy obtained from one or a few steps of Q-learning updates from $\pi$ based on data $D_0$; $R_{\pi^{\prime}}$ and $R_{\pi}$ are the actual cumulative reward of trajectories generated by policy $\pi^{\prime}$ and $\pi$, respectively, in the original RL problem. Here the meta-reward $\mathcal{R}(\pi, D_0)$ represents how much the hierarchy helps the learning progress of the local Q values' combination.

If we fix the utility network and mixing network $\pi_{\phi,\zeta}$ during meta update, then use the REINFORCE algorithm~\cite{williams1992simple}, the gradient of $\mathcal{J}_{meta}(\pi_{\theta})$ can be written w.r.t $\theta$:
\begin{equation}
    \nabla_{\theta} \mathcal{J}_{meta}=\mathbb{E}_{D_{0} \sim \pi_{\theta, \phi, \zeta}}\left[\mathcal{R}\left(\pi, D_{0}\right) \nabla_{\theta} \log \mathcal{P}\left(D_{0} \mid \pi_{\theta,\phi, \zeta}\right)\right],
\end{equation}
where $\mathcal{P}(D_0|\pi_{\theta,\phi,\zeta})$ denotes the probability of generating trajectories tuples $D_0:=\{s_t,z_t,a_t,r_t,s_{t+1}\}^T_{t=1}$given $\pi_{\theta,\phi,\zeta}$. After further derivation we can get the update rule of $\pi_{\theta}$:
\begin{equation}
    \theta \leftarrow \theta+\eta (\hat{R^{\prime}}\left(\pi, D_{1}\right) - \hat{R}\left(\pi, D_{0}\right)) \sum_{t=1}^{T} \nabla_{\theta} \log \pi_{\theta}\left(a_{t} \mid z_{t}\right),
\end{equation}

where $\hat{R}\left(\pi, D_{0}\right)$ is the cumulative episode reward sampled from $D_0$, and $\hat{R^{\prime}}\left(\pi, D_{1}\right)$ from $D_1$. $D_1$ is generated by executing new policy in the environment--a new policy from the ``exercise move'' of old $\pi_{\theta}$. .We leave the proof to the supplementary material.

For mixing network $\zeta$ after obtaining $z$ and utility network parameterized by $\phi$, they are updated by Q-learning loss. Every time before we update $\theta$, we update $\phi,\zeta$ for several times since we want the hierarchy change a bit slower than the local utility:
\begin{equation}
\begin{aligned}
    \mathcal{L}_{QL}(\phi,\zeta) &=  \mathbb{E}_{D\sim\pi_{\theta,\phi,\zeta}}[(Q_{tot}(\mathbf{a}_t,s_t;z_t,\phi, \zeta)-\\
    & [r(\mathbf{a}_t,s_t)+\gamma \max_{\mathbf{a}_{t+1}}Q_{tot}(\mathbf{a}_{t+1},s_{t+1};z_{t+1}, \phi, \zeta)])^2],\\
    \end{aligned}
\end{equation}
where $t$ is the time step, and $D$ means $D_0$ when the update of $\phi$ and $\zeta$ is for ``exercise move'' within the meta loop, and $D_{all}$(the whole replay buffer) when the update is outside.

After updating the hierarchical policy $\pi_{\theta}$, we append $D_0$ and $D_1$ to $D_{all}$ and then update $\pi_{\phi,\zeta}$ several times to keep the training-sampling ratio at 1. The data generated by the meta-update or meta-evaluation is all appended to the global replay buffer and subsequently used for Q-learning. Therefore, our method has high sample efficiency. We summarize our main algorithm in Algorithm \ref{alg:Framwork}.
\begin{algorithm}[htb] 
    \caption{Mixing Network with Meta Policy Gradient} 
    \label{alg:Framwork}
\begin{algorithmic}
     \STATE Initialize $\pi_{\theta},\pi_{\phi}, \pi_{\zeta}$
     \STATE Set learning rate $\leftarrow\eta$,meta learning rate$\leftarrow\lambda$,$D_{all}\leftarrow\{\}$
     \FOR{each episode iteration}
        % \FOR {Q-learning iteration}
        %     \STATE Generate tuple $\{s_t,z_t,a_t,r_t,s_{t+1},z_{t+1}\}^T_{t=1}$ by executing $\pi_{\theta,\phi}$
        %     \STATE $D_{all} \leftarrow D_{all} \cup \{s_t,z_t,a_t,r_t,s_{t+1},z_{t+1}\}^T_{t=1}$
        %     \STATE $\phi \leftarrow \phi + \eta \hat{\nabla}_{\phi}\mathcal{L}_{QL}(D_{all})$
        % \ENDFOR
        \STATE $D_{0}\leftarrow\{\}, D_{1}\leftarrow\{\}$ 
        \STATE Generate tuple $\{s_t,z_t,a_t,r_t,s_{t+1},z_{t+1}\}^T_{t=1}$ by executing $\pi_{\theta,\phi,\zeta}$
        \STATE $D_{0} \leftarrow D_{0} \cup \{s_t,z_t,a_t,r_t,s_{t+1},z_{t+1}\}^T_{t=1}$
        \STATE $R_0\leftarrow \sum^T_{t=1}r_t$
        \FOR{exercise move iteration}
            \STATE $\phi \leftarrow \phi + \eta \hat{\nabla}_{\phi}\mathcal{L}_{QL}(D_{0})$
            \STATE $\zeta \leftarrow \zeta + \eta \hat{\nabla}_{\zeta}\mathcal{L}_{QL}(D_{0})$
        \ENDFOR
        \STATE Generate tuple $\{s_t,z_t,a_t,r_t,s_{t+1},z_{t+1}\}^T_{t=1}$ by executing $\pi_{\theta,\phi,\zeta}$
        \STATE $D_{1} \leftarrow D_{1} \cup \{s_t,z_t,a_t,r_t,s_{t+1},z_{t+1}\}^T_{t=1}$
        \STATE $R_1\leftarrow \sum^T_{t=1}r_t$
        \STATE $\mathcal{R}\leftarrow R_1-R_0$
        \STATE $\theta \leftarrow \theta +\lambda\hat{\nabla}_{\theta}\mathcal{J}_{meta}(D_0,\mathcal{R})$
        \STATE $D_{all}\leftarrow D_{all}\cup D_0 \cup D_1$
        \FOR {Q-learning iteration}
            \STATE $\phi \leftarrow \phi + \eta \hat{\nabla}_{\phi}\mathcal{L}_{QL}(D_{all})$
            \STATE $\zeta \leftarrow \zeta + \eta \hat{\nabla}_{\zeta}\mathcal{L}_{QL}(D_{all})$
        \ENDFOR
     \ENDFOR
\end{algorithmic}
\end{algorithm}

\begin{figure*}[!htbp]
\begin{center}
\includegraphics[width=\textwidth]{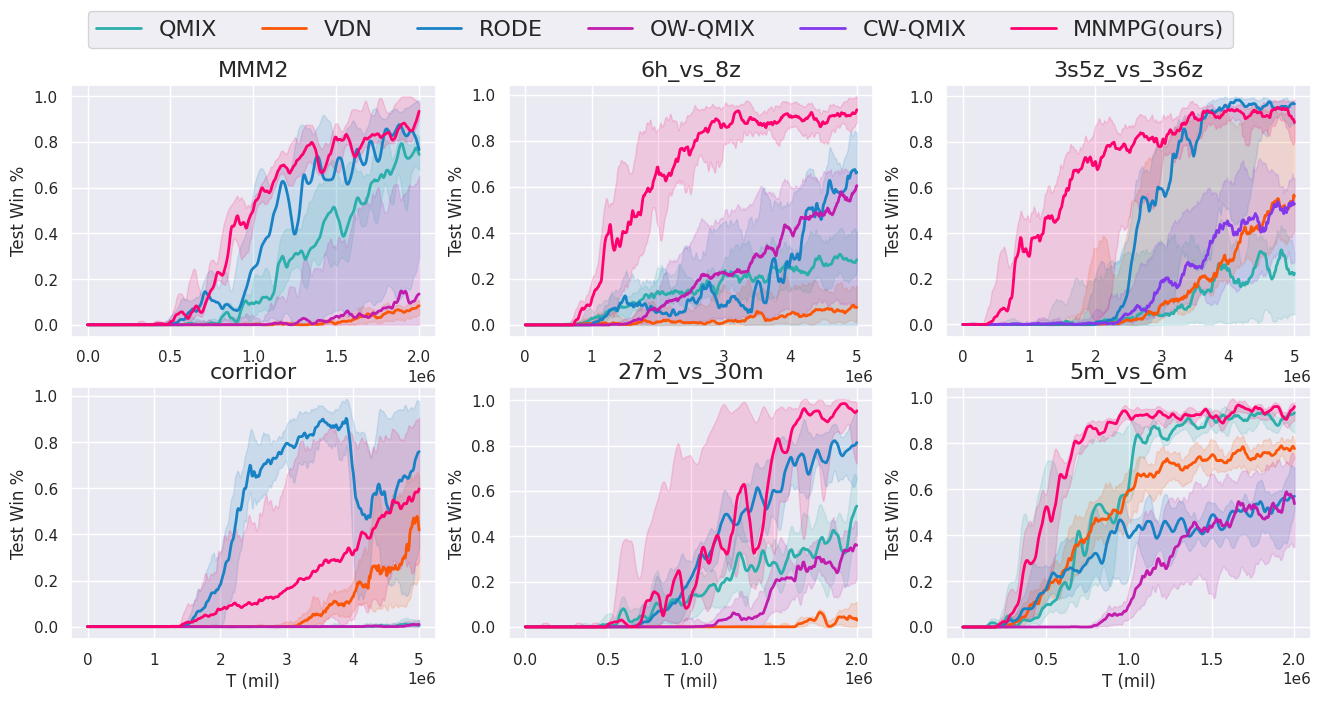}
\end{center}
\caption{Results on all 5 super hard maps and 1 hard map(\texttt{5m\_vs\_6m}) of SMAC.}
\label{rlp}
\end{figure*}

\subsection{Strong Adaptation of MNMPG}
MNMPG is the improvement of the traditional Q mixing network that satisfies the monotonicity constraint between $Q_{tot}$ and each $Q_i$ produced by the local utility network:
\begin{equation}
    \frac{\partial Q_{t o t}}{\partial Q_{i}} \geq 0, \forall i \in I.
\end{equation}
The monotonicity constraint ensures that the $\operatorname{argmax}$ performed on $Q_{tot}$ yields the same result as a set of individual $\operatorname{argmax}$ operations performed on each $Q_i$~\cite{rashid2018qmix}:
\begin{equation}
    \underset{\mathbf{a}}{\operatorname{argmax}} Q_{\operatorname{tot}}(\boldsymbol{\tau}, \mathbf{u})=\left(\begin{array}{c}
\operatorname{argmax}_{a^{1}} Q_{1}\left(\tau^{1}, a^{1}\right) \\
\vdots \\
\operatorname{argmax}_{a^{n}} Q_{n}\left(\tau^{n}, a^{n}\right)
\end{array}\right),
\end{equation}
which is the theoretical guarantee for the optimality of CTDE. MNMPG inherits such monotonicity constraint by taking the absolute value of the vector decoded from the global hierarchy before it is element-wisely multiplied  to the local $Q_i$. Since the hierarchy is calculated from the full state, and weights of the local $Q_i$ can condition on the global hierarchy arbitrarily, we can integrate the full information for centralized training flexibly and globally. 

In principle, our method can be integrated into any method that utilizes a monotonic mixing network by replacing the corresponding mixing network with MNMPG. Role-based methods such as MAVEN~\cite{mahajan2019maven}, ROMA~\cite{wang2020roma}, RODE~\cite{wang2020rode} all allocate a role or a mode of exploration for each local utility network to enhance its ability to deal with local observations, while their global hierarchy is the simple dependency on full state~(and a fixed noise during each episode for MAVEN). We implement our structure on the mixing network of RODE, adding global guidance on the combination of $Q_i$. In experiments, higher performance and better stability are observed.

\section{Experiments}
In this section, we carry out experiments on the StarCraft II micromanagement (SMAC) benchmark~\cite{samvelyan2019starcraft}, a mainstream benchmark for research on CTDE algorithms. SMAC has a variety of maps with high learning complexity, which is suitable for testing our method on various collaborative multi-agent situations. SMAC has continuous observation space and discrete action space. Depending on the map difficulty, the dimension of action space ranges from 9 $\sim$ 36, which is relatively large. Meanings of action include \texttt{move} towards four cardinal directions, \texttt{stop}, take \texttt{noop}(do nothing), and select an enemy/ally to \texttt{attack/heal}.

We show in Sec.~\ref{sec:perform} the overall performance of MNMPG. Then in Sec.~\ref{sec:gde}, we explore how a hierarchical mixing network affects the joint exploration procedure. Moreover, in Sec.~\ref{sec:scale}, we show the results of implementing our method on other existing CTDE methods. Finally, in Sec.~\ref{sec:ablation}, we show a detailed discussion on the construction and parameters of MNMPG.

We perform the experiments on 2 servers with 3 RTX-2080TI-11G GPU each and a server with 3 RTX-3090-24G GPU, accompanied with 2 Intel(R) Core(TM) i9-9960X CPU @ 3.10GHz and one Intel(R) Xeon(R) Gold 6248R CPU @ 3.00GHz individually. 
If not specifically mentioned, the optimizer is RMSprop with a learning rate of $5\times10^{-4}$, $\alpha$ of $0.99$, and with no momentum or weight decay for all experiments. The exploration strategy is $\epsilon$-greedy with $\epsilon$ annealed linearly from $1.0$ to $0.05$, while the anneal time depends on the difficulty of the map based on the discovery from \cite{rashid2020weighted}. At every training procedure, a batch of 32 episodes is sampled from the replay buffer. Every 20000 environment steps, the policy will be tested for 24 episodes, and the winning rate against the preconfigured AI will be recorded. A $5\times10^6$ time steps' experiment takes about 27 hours to run. 
% We leave the detailed experiment settings and hyperparameters to the supplementary material. 
For all baseline algorithms, we use the public codes provided by the author, while the hyperparameters are selected from the corresponding paper.

\begin{figure*}[!htbp]
\begin{center}
\includegraphics[width=\textwidth]{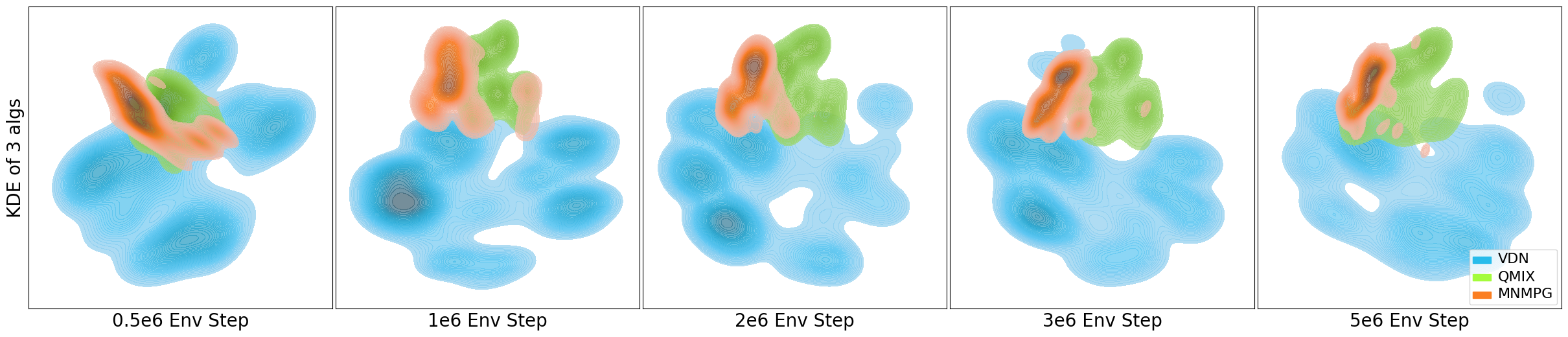}
\end{center}
\caption{The kernel density estimate(KDE) map of \texttt{6h\_vs\_8z}'s state visitation after t-SNE for 3 algorithms.}
\label{kde}
\end{figure*}

\subsection{Performance}\label{sec:perform}
We show in Fig.~\ref{rlp} the test winning rate curve of 5 algorithms relative to the environment steps during training on all 5 \textbf{super hard} maps and one \textbf{hard} map$(\texttt{5m\_vs\_6m})$. More resutls on \textbf{hard} and \textbf{easy} maps are appended in the supplementary material. Algorithms include QMIX~\cite{rashid2018qmix}, VDN~\cite{sunehag2018value}, RODE~\cite{wang2020rode}, OW/CW-QMIX~\cite{rashid2020weighted}, and MNMPG(ours), in which RODE is the current state-of-the-art on super hard SMAC maps. For map \texttt{3s5z\_vs\_3s6z}, we use CW-QMIX as the baseline since its performance is better than OW-QMIX in ~\cite{rashid2020weighted}, and for other maps, we use OW-QMIX instead.  We set $\epsilon$ anneal time to $5\times10^5$ environment steps for map \texttt{27m\_vs\_30m}, \texttt{6h\_vs\_8z}, and \texttt{3s5z\_vs\_3s6z}. Others have $\epsilon$ anneal time $5\times10^4$. Each curve is drawn from 5 random seeds. The solid line represents the medium test winning rate, while the background shadow is the maximum and minimum value. 

We can infer from the figure that MNMPG exceeds the current state-of-the-art on 4 of 5 super hard maps, as well as hard map \texttt{5m\_vs\_6m}. Furthermore, on \texttt{6h\_vs\_8z} and \texttt{3s5z\_vs\_3s6z}, MNMPG has a  distinguished performance. But on map \texttt{corridor}, it has difficulty to learn a great model in $5\times 10^6$ environment steps. We notice that our local utility network is a simple 2-layer MLP, similar to QMIX, and QMIX performs worse on \texttt{corridor}. It can not win even once after the whole 5e6 time steps' training. Therefore, a possible reason for bad performance is that the capacity of 2-layer MLP is so short to handle the complex observation on \texttt{corridor}, since in \texttt{corridor} the positions of each side are diagonal, and agents can not move toward a certain direction to get close to the enemy. Agents must learn to move east and north alternately. Another perspective is the relationship between MNMPG and QMIX. MNMPG can be regarded as a modification to QMIX, which leads the model to explore more on states that may result in reward improvement. Nevertheless, if all visited states produce low rewards like the performance of QMIX in the map \texttt{corridor}, the meta mechanism can not invent a new way to improve the reward. It is more like an accelerator than an explorer. We replace the utility network with that from RODE and as shown in Sec.\ref{sec:scale} our method performs better on the map \texttt{corridor}.

\subsection{Guided Exploration}\label{sec:gde}
To study the importance of our meta policy, we run three algorithms on map \texttt{6h\_vs\_8z} with the same parameters: QMIX, VDN, and our MNMPG. These three algorithms are only different in mixing networks. VDN adds up the $Q$ values from the local utility network as $Q_{tot}$. QMIX utilizes a 2-layer weighted sum of local $Q$ values, where the weights are computed directly from the full states. MNMPG has a similar structure to QMIX, while the weights of local $Q$ values are computed from a global hierarchy. We test each algorithm 32 times every $5\times10^5$ environment steps and save the whole trajectories to plot the state visitation map. To visualize the trajectories with the high dimensional state, we use t-SNE~\cite{maaten2008visualizing} to reduce each state's dimension to 2 and plot the state visitation map in Fig.~\ref{kde}. The color of state visitation is calculated by kernel density estimate. The darker the color, the higher the frequency of state visitation. Each column is the different environment steps during training, representing different stages of the training procedure. Since the parameter $\epsilon$ of $\epsilon$-greedy exploration strategy starts keeping unchanged at $5\times10^5$ environment steps, we omit the previous state for its high probability of random exploration. We keep the same scale and location for the three algorithms, so we hide the axis labels. 

We can infer from Fig.~\ref{kde} that MNMPG focuses more on exploring specific areas, while QMIX tends to explore a relatively large area, and VDN explores most of the state space almost uniformly. Meanwhile, MNMPG has a more concentrated exploration area with the learning progress, while QMIX and VDN have no such phenomenon. The state visitation map can explain our method's better performance: Compared with QMIX's random exploration, the hierarchy with meta policy gradient gives clear guidance on exploration, which is easy to generate a better reward. Moreover, the higher reward signal in turn gives better guidance by meta-learning.

\begin{figure}[!htbp]
\begin{center}
\includegraphics[width=0.45\textwidth]{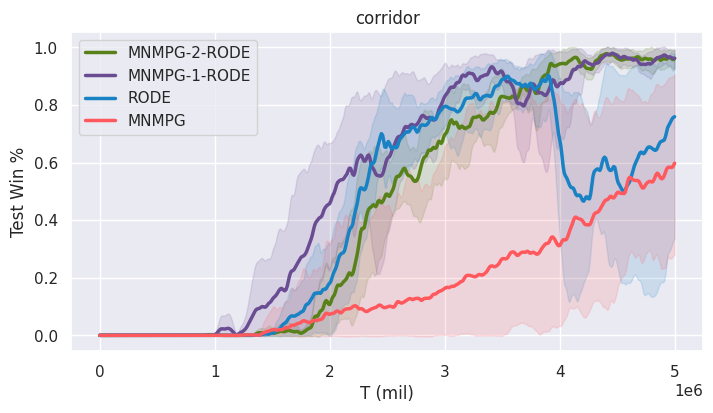}
\end{center}
\caption{Adaptation to RODE.}
\label{adaptation}
\end{figure}
\subsection{Adaptation to Other Methods}\label{sec:scale}
To test our method's adaptation, we modify the mixing network of RODE~\cite{wang2020rode}, which is the current state-of-the-art on super hard SMAC maps. The original RODE has a bi-level utility network. The role selector searches in a smaller role space and lower temporal resolution, while role policies learn
in significantly reduced primitive action-observation spaces. The role selector and role policies have an individual mixing network separately. We implement two kinds of modification on mixing network: Utilizing MNMPG for role policies merely or learning two MNMPG for role selector and role policies separately. We carry out experiments on map \texttt{corridor}, on which our method with a simple utility network has poor performance. The results are shown in Fig.~\ref{adaptation}. Legend "MNMPG-1-RODE" represents the MNMPG modification on role policies only, and legend "MNMPG-2-RODE" is the MNMPG modification on both levels. Legend "MNMPG" means the MNMPG with a simple two-layer utility network. We repeat each experiment for 5 seeds.

The results demonstrate that the original RODE is unstable during training. When close to defeating all enemies, the performance of RODE drops sharply. Replacing the original mixing network with MNMPG helps RODE stabilize the training procedure and improve the performance. Furthermore, MNMPG for role policies performs slightly better than MNMPG for both role selector and role policies. We speculate that in RODE, the role selector plays as a hierarchical information processor. Therefore, no extra global hierarchy is needed. However, the role policies concentrate on dealing with local information, so a clear global hierarchy can apparently improve its ability.

\begin{figure*}[!htbp]
\begin{center}
\includegraphics[width=\textwidth]{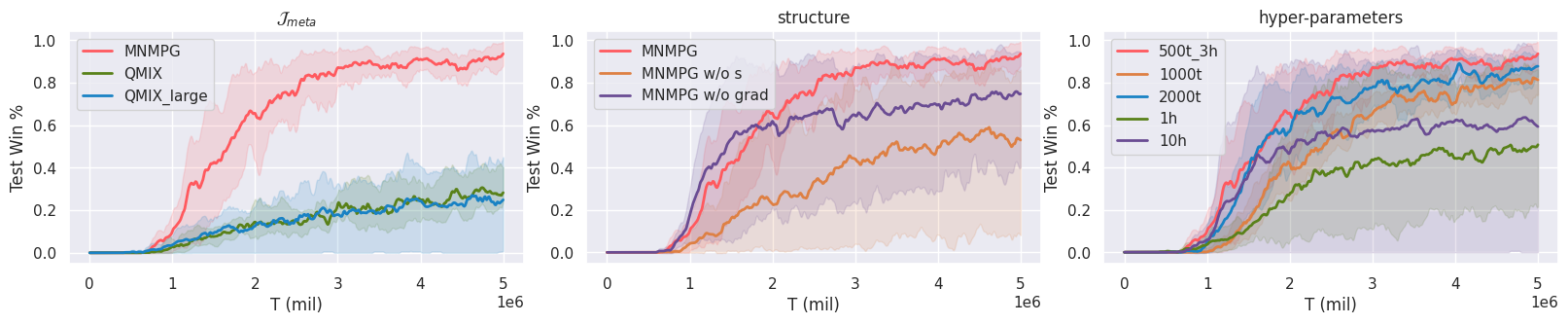}
\end{center}
\text{\ \ \ \ \ \ \ \ \ \ \ \ \ \ \ \  \ \ \ \ \ \ \ \ \ \ \ \ \ \ \ \ \ \   (a)\ \ \ \ \ \ \ \ \ \ \ \ \ \ \ \ \ \ \ \ \  \ \ \ \ \ \ \ \ \ \ \ \ \ \ \ \  \ \ \ \ \ \ \ \ \ \ \ \ \ \ \ \ \ \ \ \ \ \ (b)\ \ \ \ \ \ \ \ \ \ \ \ \ \ \ \ \ \ \ \ \ \ \ \ \ \ \ \ \ \ \ \ \ \ \ \ \ \ \ \ \ \ \ \ \ \ \ \ \ \ \ \ \ \ \ \ \ \ \ \ \ \ (c)}
\caption{Ablation study of MNMPG.}
\label{ablation}
\end{figure*}

\subsection{Ablation Study}\label{sec:ablation}
We carry out several ablation studies on the map \texttt{6h\_vs\_8z} to determine the importance of each component of MNMPG, shown in Fig.~\ref{ablation}. Each experiment is repeated for three random seeds. The solid line represents the average test winning rate, while the background shadow is the maximum and minimum value. 

We first consider setting the learning rate of $\mathcal{J}_{meta}$ to $0$, then HPMG degenerates into QMIX with a more extensive mixing network whose weights of Q values are computed from an encoder-decoder-like structure. We can infer from Fig.~\ref{ablation}(a) that merely increasing the network's scale~(corresponding to the legend ``QMIX\_large'') can not bring any performance improvement. That shows the importance of the hierarchical guidance of meta policy gradient.

Then we consider the modification of network structure. In MNMPG, we allow the gradient flow of $\mathcal{L}_{QL}$ to update the hierarchical policy $\pi_{\theta}$, and we concatenate the original full state to the vector decoded from the global hierarchy as the input of the mixing network. In Fig.~\ref{ablation}(b), for legend ``MNMPG w/o grad'' we block the gradient from $\mathcal{L}_{QL}$, leaving only meta policy gradient to update it. For legend ``MNMPG w/o s'', the vector without the full state is used as the mixing network input. We can infer from the figure that blocking the gradient harm the winning rate by about 15\%, which may come from the instability of the policy gradient. We find that if the Q-learning update is interspersed between meta policy gradient, the variance of policy gradient will be smaller. Meanwhile, concatenating the full state allows the mixing network to not only combine the local Q values from the global hierarchy but also from the local details that may be omitted, therefore improving the overall performance for around 40\%.

Finally, we discuss the hyperparameters, including meta update interval and dimension of global hierarchy . We set the meta update interval in MNMPG to 500 environment steps and set the global hierarchy dimension to 3. In Fig.~\ref{ablation}(c), legend ``1000t'' and ``2000t'' show the results when decreasing the meta update frequency to 0.5 and 0.25. They both perform worse than the original MNMPG, whose legend is ``500t\_3h''. And we show in Fig.~\ref{ablation}(c) legends ``1h'' and ``10h'' the results with dimension 1 and 10 for global hierarchy, respectively. They both have worse performance than dimension 3, and dimension 10 performs relatively better. We can infer that a one-dimensional global hierarchy can not offer enough information to combine local Q values, while dimension 10 is too large for a simple and overarching global hierarchy, distinguished from the full state.

\section{Related Works}
\textbf{Collaborative Multi-agent Reinforcement Learning}. Recently, multi-agent reinforcement learning has shown strong performance in games~\cite{berner2019dota,samvelyan2019starcraft,ye2020towards,jaderberg2019human} and real-world phenomena such as the emergence of tool usage~\cite{baker2019emergent}, communication~\cite{foerster2016learning,sukhbaatar2016learning,ding2020learning,tian2020joint}, and social influence~\cite{jaques2019social}. We mainly focus on the typical paradigm of centralized training with decentralized execution~(CTDE) for collaborative multi-agent tasks in this part. 

In \cite{tampuu2017multiagent}, Independent Q-learning~\cite{tan1993multi} is used to train DQN~\cite{mnih2015human} for each agent. It treats other agents as part of the environment, which causes instability during training~\cite{claus1998dynamics,foerster2017stabilising}. Another way is joint action learning~\cite{claus1998dynamics} that regards all agents as the whole and utilizes single-agent RL algorithms to solve it, ignoring the internal relations between agents and has difficulty distributing. A trade-off that combines the advantages of independent Q-learning and joint action learning is the paradigm of centralized training with decentralized execution(CTDE).

Many works concentrate on how to combine the Q values from the local utility network with the CTED paradigm. VDN~\cite{sunehag2018value} uses Q-learning on $Q_{tot}$, which is the sum of local $Q_i$, to ensure that the global $\operatorname{argmax}$ on $Q_{tot}$ yields the local $\operatorname{argmax}$ on each agent $Q_i$, which is called the Individual-Global-Max (IGM) principle. QMIX~\cite{rashid2018qmix} relaxes the sufficient condition of the IGM principle to $\frac{\partial Q_{t o t}}{\partial Q_{i}} \geq 0$, and further modifies the mixing network from a simple sum operator to a neural network with non-negative weights condition on full state $s$. Qatten~\cite{yang2020qatten} combines an attention-based mechanism to the weight of the first layer of QMIX. Other similar architecture includes \cite{bohmer2019exploration} and \cite{wen2020smix}. Furthermore, QPLEX~\cite{wang2020qplex} takes a duplex dueling network architecture to factorize the joint value function. QTRAN~\cite{son2019qtran} is another Q-learning-based algorithm that learns an unrestricted joint action Q function to loosen the bondage of the non-negative weights constraint. However, its empirical performance is not satisfactory on complex tasks. Weighted QMIX~\cite{rashid2020weighted} learns another mixer that does not satisfy the IGM principle to adjust samples' weight for the original QMIX. Our method can be concluded as another simple but efficient and effective architecture of mixing networks that satisfies the IGM principle. 

There are other works on learning a local utility network with powerful capacity. MAVEN~\cite{mahajan2019maven} allocate a hierarchical mode of exploration to each local agent, avoiding the failure of QMIX on the specific situation if each joint action is visited uniformly. ROMA~\cite{wang2020roma} claims to learn identifiable and specialized roles for each local agent. RODE~\cite{wang2020rode} uses a bi-level structure to learn roles that are highly interpretable. It firstly splits action spaces for different roles, then learns q functions in the shrunken action space. Actor-critic is another branch of CTDE. MADDPG~\cite{lowe2017multi} trains local policy via the multi-agent deterministic policy gradient theorem. COMA~\cite{foerster2018counterfactual} uses a counterfactual baseline for training the centralized critic. Moreover, LICA~\cite{zhou2020learning} directly uses local agents' policy as the input of the centralized critic, isolating the state transformation from the action transformation. The above methods all have a simple mixing network or a centralized critic. Therefore, they have the potential to be extended by our method.

\textbf{Meta Reinforcement Learning} Some methods are using the meta-objective (usually the difference of episode return ) for reinforcement learning. Meta-knowledge is used to construct loss~\cite{zhou2020online,veeriah2019discovery}, reward~\cite{jaderberg2019human,zheng2018learning}, or hyperparameters~\cite{xu2018meta,young2018metatrace}. In \cite{xu2018learning} a teacher-student strategy is proposed for global exploration.

\section{Conclusion}
This paper presents Mixing Network with Meta Policy Gradient~(MNMPG), a framework for collaborative multi-agent reinforcement learning with the CTDE paradigm. We find that a well-designed mixing network with a simple local utility network can produce outstanding performance on challenging collaborative multi-agent tasks. Specifically, we train a global hierarchy with meta policy gradient to assign proper credit to each local agent. The hierarchy is entirely based on self-improvement, hence using no prior information. With a simple local utility network, we achieve state-of-the-art on most of the super hard SMAC maps. When combined with a role-based local utility network, higher performance and better stability are observed.

% Since our method uses policy gradient for learning global hierarchy, we suffer from its instability. It can be inferred from the experimental results that our method has a relatively large variance in the test winning rate. In the future, we will work on stabilizing the training procedure and try to introduce the actor-critic based methods.

\bibliography{arxiv}
\bibliographystyle{icml2021}

\newpage
\renewcommand\thefigure{\thesection.\arabic{figure}}
\renewcommand\thetable{\thesection.\arabic{table}}
\setcounter{figure}{0}
\setcounter{table}{0}
\appendix

\section{Proof of the Meta-Policy Gradient}
The meta objective is the reward difference between before and after several ``exercise move'':
\begin{equation}
    \begin{aligned}
    \mathcal{J}_{meta}(\pi_{\theta}) &=\mathbb{E}_{D_{0} \sim \pi_{\theta,\phi,\zeta}}\left[\mathcal{R}\left(\pi, D_{0}\right)\right] \\ &=\mathbb{E}_{D_{0} \sim \pi_{\theta,\phi,\zeta}}\left[R_{\pi^{\prime}}-R_{\pi}\right],
    \end{aligned}
\end{equation}
Use the REINFORCE~\cite{williams1992simple} trick, we can write the meta objective into the derivative of $\log \mathcal{P}$:
\begin{equation}
    \nabla_{\theta} \mathcal{J}_{meta}=\mathbb{E}_{D_{0} \sim \pi_{\theta, \phi, \zeta}}\left[\mathcal{R}\left(\pi, D_{0}\right) \nabla_{\theta} \log \mathcal{P}\left(D_{0} \mid \pi_{\theta,\phi, \zeta}\right)\right],
\end{equation}
Then factorize the distribution $\mathcal{P}$:
\begin{equation}
    \mathcal{P}\left(D_{0} \mid \pi_{\theta,\phi, \zeta}\right) = p(s_0)\prod^T_{t=1}\pi_{\theta,\phi, \zeta}(a_t|s_t)p(s_{t+1}|s_t,a_t),
\end{equation}
where $p(s_0)$ is the initial distribution of the environment and $p(s_{t+1}|s_t,a_t)$ is the transition probability. $p(s_0)$ and $p(s_{t+1}|s_t,a_t)$ have no dependency on the policy $\pi_{\theta,\phi, \zeta}$. Moreover, the mixing network $f_{\zeta}$ and the utility network $f_{\phi}$ are fixed during meta-update. Therefore, the update rule for meta policy $\pi_{\theta}$ can be written as:
\begin{equation}
\begin{aligned}
    \nabla_{\theta} \mathcal{J}_{meta}=&\mathbb{E}_{D_{0} \sim \pi_{\theta, \phi, \zeta}}  (\mathcal{R}  \left(\pi, D_{0}\right) \nabla_{\theta} \log [p(s_0)  \prod^T_{t=1}\pi_{\theta,\phi, \zeta}(a_t|s_t)p(s_{t+1}|s_t,a_t)])\\
    =&\mathbb{E}_{D_{0}\sim \pi_{\theta, \phi, \zeta}} (\mathcal{R}  \left(\pi, D_{0}\right) \nabla_{\theta}[\sum^T_{t=1} \log \pi_{\theta, \phi, \zeta}(a_t|s_t)]+
    \nabla_{\theta}[\log p(s_0)  +\sum^T_{t=1}\log p(s_{t+1}|s_t,a_t)])\\
    =&\mathbb{E}_{D_{0}\sim \pi_{\theta, \phi, \zeta}} (\mathcal{R}  \left(\pi, D_{0}\right) \nabla_{\theta}[\sum^T_{t=1} \log \pi_{\theta, \phi, \zeta}(a_t|s_t)]\\
    =&\mathbb{E}_{D_{0}\sim \pi_{\theta, \phi, \zeta}} ([R' \left(\pi, D_{1}\right) - R\left(\pi, D_{0}\right)] 
    \nabla_{\theta}[\sum^T_{t=1}  \log \pi_{\theta, \phi, \zeta}(a_t|s_t)])
\end{aligned}
\end{equation}

\section{More Results}
\begin{figure*}[!htbp]
\begin{center}
\includegraphics[width=0.6\textwidth]{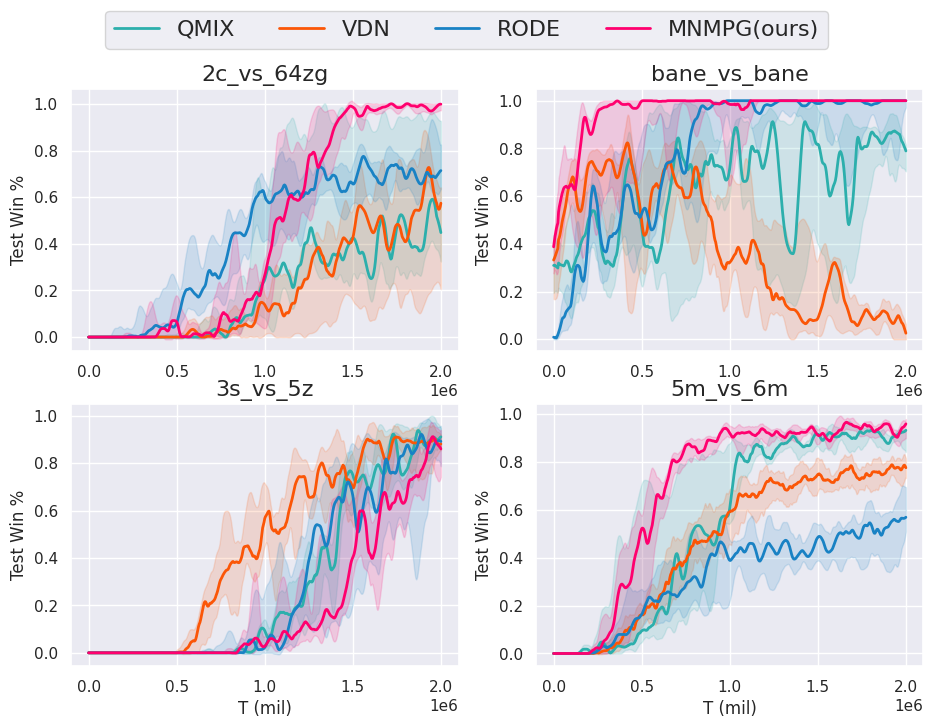}
\end{center}
\caption{Results on all hard maps.}
\label{hard}
\end{figure*}

\begin{figure*}[!htbp]
\begin{center}
\includegraphics[width=0.6\textwidth]{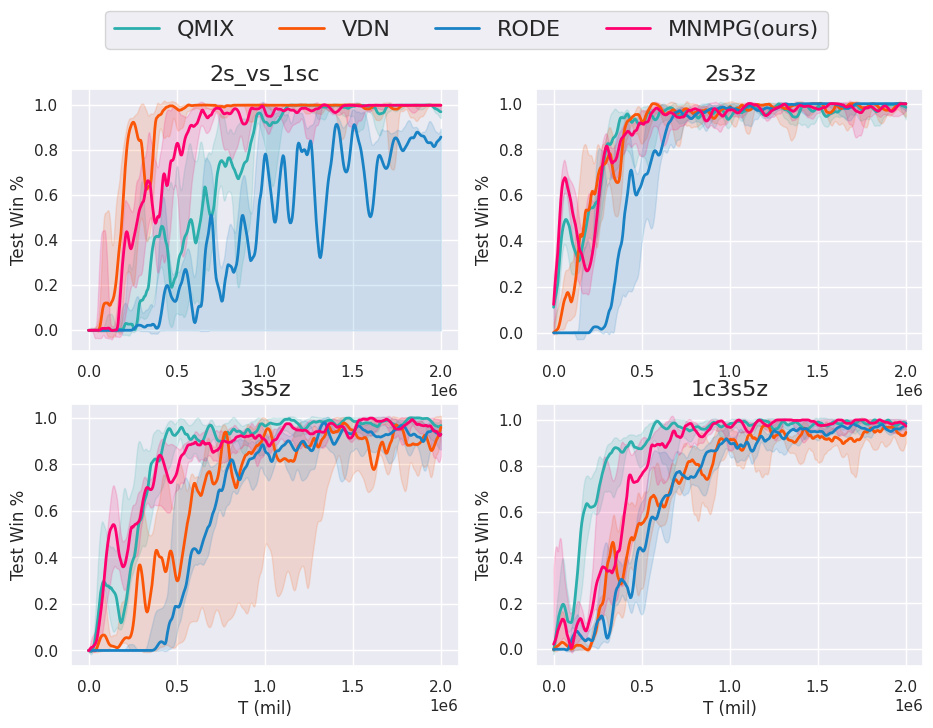}
\end{center}
\caption{Results on 4 easy maps.}
\label{easy}
\end{figure*}

Except for the maps mentioned in the text, we evaluate MNMPG on 4 easy and all 4 hard SMAC maps. We can infer from Fig.\ref{hard} that MNMPG exceeds the current state-of-the-art on 3 of 4 hard maps. And in Fig.\ref{easy} we show that MNMPG has similar performance to the current state-of-the-art on easy maps.

\end{document}